\documentclass{llncs}
\usepackage{makeidx}  % allows for indexgeneration
\usepackage{amsmath}
\usepackage{amssymb}
\usepackage{algorithm}   % pseudocode
\usepackage{algorithmic} % pseudocode

\usepackage[pdftex]{graphicx}

\begin{document}
\frontmatter          % for the preliminaries
\pagestyle{headings}  % switches on printing of running heads
\addtocmark{ARC Paper Content} % additional mark in the TOC
\title{Approximate FPGA-based LSTMs under Computation Time Constraints}
\titlerunning{Approximate FPGA-based LSTMs under Computation Time Constraints}  % abbreviated title (for running head)
\author{Michalis~Rizakis \and
Stylianos~I.~Venieris \and
Alexandros~Kouris \and \\
Christos-Savvas Bouganis}
\authorrunning{Michalis Rizakis et al.} % abbreviated author list (for running head)
\tocauthor{Michalis Rizakis, Stylianos I. Venieris, Alexandros Kouris, and Christos-Savvas Bouganis}
\institute{Dept. of Electrical and Electronic Engineering, Imperial College London\\
\email{\{michail.rizakis14, stylianos.venieris10, a.kouris16, christos-savvas.bouganis\}@imperial.ac.uk} 
}

\maketitle              % typeset the title of the contribution

\vspace{-0.7cm}
\begin{abstract}
Recurrent Neural Networks and in particular Long Short-Term Memory (LSTM) networks have demonstrated state-of-the-art accuracy in several emerging Artificial Intelligence tasks. However, the models are becoming increasingly demanding in terms of computational and memory load. Emerging latency-sensitive applications including mobile robots and autonomous vehicles often operate under stringent computation time constraints. In this paper, we address the challenge of deploying computationally demanding LSTMs at a constrained time budget by introducing an approximate computing scheme that combines iterative low-rank compression and pruning, along with a novel FPGA-based LSTM architecture. Combined in an end-to-end framework, the approximation method's parameters are optimised and the architecture is configured to address the problem of high-performance LSTM execution in time-constrained applications. Quantitative evaluation on a real-life image captioning application indicates that the proposed methods required up to 6.5$\times$ less time to achieve the same application-level accuracy compared to a baseline method, while achieving an average of 25$\times$ higher accuracy under the same computation time constraints.

% The state-of-the-art performance achieved by Deep Neural Networks on many emerging applications has increasingly gained interest of many communities to employ DNNs on a wide span of tasks. When long-range dependencies exists on input data, Recurrent NNs, and especially LSTMs, have demonstrated great learning capacity, which however comes with high computational and memory demands. In this paper, we address these challenges by introducing an approximate computing scheme that combines iterative low rank compression and prunning, along with a novel reconfigurable hardware architecture for LSTMs on FPGAs. Combined on an end-to-end framework, approximation method's and architecture's parameters are co-optimised to address the problem of high perforance LSTM execution in time-constrained applications by a tailored architecture. Examining image captioning as a case study the proposed method required up to 6.51x less time to achieve the same application-level accuracy when compared to a baseline method, while achiving 24.88x higher accuracy on average under computation time constraints. 
\vspace{-0.2cm}
\end{abstract}
\vspace{-1.0cm}

\section{Introduction}
\vspace{-0.25cm}
%
% For the introduction I want to include the reasoning and motivation behind our work. Meaning, why we need to reduce the computational burden of an RNN, and more specifically why we chose to apply our methods on an LSTM. In order to achieve this, I would first start by giving a brief description of RNNs, with the focus on its parameters (weights), the computations that need to happen, why they can be sometimes too many for a simple FPGA, and how we can reduce them in order for the inference stage of an RNN to run on an FPGA. Afterwards, I would go on to analyze the reasoning behind choosing LSTM as the target RNN, as well as why we chose image captioning as our application.

Recurrent Neural Networks (RNNs) is a machine learning model which offers the capability of recognising long-range dependencies in sequential and temporal data. RNN models, with the prevalence of Long Short-Term Memory (LSTMs) networks, have demonstrated state-of-the-art performance in various %emerging
AI applications including % machine translation \cite{Sutskever_2014},
scene labelling \cite{byeon2015scene} and image generation \cite{gregor15}. Moreover, LSTMs have been successfully employed for AI tasks in complex environments including human trajectory prediction \cite{Alahi_2016_CVPR} and ground classification \cite{otte2016recurrent} on mobile robots, with more recent systems combining language and image processing in tasks such as image captioning \cite{Vinyals_2017} and video understanding \cite{Donahue_2017}.

Despite the high predictive power of LSTMs, their computational and memory demands pose a challenge with respect to deployment in latency-sensitive and power-constrained applications. Modern intelligent systems such as mobile robots and drones that employ LSTMs to perceive their surroundings often operate under time-constrained, latency-sensitive settings. In such scenarios, retrieving the best possible output from an LSTM given a constraint in computation time may be necessary to ensure the timely operation of the system. Moreover, the requirements of such applications for low absolute power consumption, which would enable a longer battery life, %and flight time,
prohibit the deployment of high-performance, but power-hungry platforms, such as multi-core CPUs and GPUs. In this context, reconfigurable computing in the form of Field-Programmable Gate Arrays (FPGAs) constitutes a promising target device that can combine customisation and reconfigurability to achieve high performance at a low power envelope.

% \cite{Denil_2013}

In this work, an approximate computing scheme along with a novel hardware architecture for LSTMs are proposed as an end-to-end framework to address the problem of high-performance LSTM execution in time-constrained settings. Our approach proposes an iterative approximation method that applies simultaneously low-rank compression and pruning of the LSTM model with a tunable number of refinement iterations. The iterative property enables our framework to (i) exploit the resilience of the target application to approximations, (ii) explore the trade-off between computational and memory load and application-level accuracy and (iii) run the LSTM under a time constraint with increasing accuracy as a function of computation time budget. At the hardware level, our system consists of a novel reconfigurable architecture mapped on an FPGA which exploits the inherent parallelism of the LSTM, parametrised with respect to the level of compression and pruning. By optimising the parameters of the approximation method, the proposed framework generates a system tailored to the target application, the available FPGA resources and the computation time constraints.

\vspace{-0.5cm}

\section{Background}
%I am still unsure about the title of this section, but I want to include a brief introduction of LSTMs as well as Singular Value Decomposition, in order to introduce the mathematical context needed for the explanation of the method. The way I want this section to be organized is in two subsections.
% * <a.kouris16@imperial.ac.uk> 2017-11-07T19:51:21.809Z:
% 
% The LSTM part should probably include information about the structure of LSTMs, along with a formal analysis of the computations performed in the main layers, to form a basis on describing where/how the SVD-based approximations fit in the next chapter. The SVD is considered a well-known linear algebra transformation and I would not dedicate a subsection to it. 
% 
% ^.
\vspace{-0.2cm}

\subsection{LSTM Model}
\vspace{-0.2cm}
A vanilla RNN typically processes an input and generates an output at each time step. Internally, the network has recurrent connections from the output at one time step to the hidden units at the next time step which enables it to capture sequential patterns.
%The Long Short-Term Memory (LSTM) model is a type of RNN that has been successfully used in a variety of applications. 
The LSTM model differs from vanilla RNNs in that it comprises control units named gates, instead of layers. A typical LSTM has four gates. %, each one of which serving a specific purpose. 
The \textit{input} gate (Eq. (\ref{lstm_eq1})), along with the \textit{cell} gate (Eq. (\ref{lstm_eq4})) are responsible for determining how much of the current input will propagate to the output. The \textit{forget} gate (Eq. (\ref{lstm_eq2})) is responsible for determining whether the previous state of the LSTM will be forgotten or not, while the \textit{output} gate (Eq. (\ref{lstm_eq3})) determines how much of the current state will be allowed to propagate to the final output of the LSTM at the current time step. Computationally, the gates are matrix-vector multiplication blocks, followed by a nonlinear elementwise activation function. The equations for the LSTM model are shown below:
\vspace{-0.2cm}
\begin{eqnarray}
\label{lstm_eq1}
\vec i^{(t)}&=&\sigma(\boldsymbol{W}_{ix}\vec x^{(t)}+\boldsymbol{W}_{ih}\vec h^{(t-1)})\\
\label{lstm_eq2}
\vec f^{(t)}&=&\sigma(\boldsymbol{W}_{fx}\vec x^{(t)}+\boldsymbol{W}_{fh}\vec h^{(t-1)})\\
\label{lstm_eq3}
\vec o^{(t)}&=&\sigma(\boldsymbol{W}_{ox}\vec x^{(t)}+\boldsymbol{W}_{oh}\vec h^{(t-1)})\\
\label{lstm_eq4}
\vec c^{(t)}&=&\vec f^{(t)}\odot \vec c^{(t-1)}+\vec i^{(t)} \odot tanh(\boldsymbol{W}_{cx}\vec x+\boldsymbol{W}_{ch}\vec h^{(t-1)}) \\
\label{lstm_eq5}
\vec h^{(t)}&=&\vec c^{(t)} \odot \vec o^{(t)}
\vspace{-0.3cm}
\end{eqnarray}
% \vspace{-0.2cm}
$\boldsymbol{i}^{(t)}, \boldsymbol{f}^{(t)}$ and $\boldsymbol{o}^{(t)}$ are the \textit{input}, \textit{forget} and \textit{output} gates respectively, $\boldsymbol{c}^{(t)}$ is the current state of the LSTM, $\vec h^{(t-1)}$ is the previous output, $\vec x^{(t)}$ is the current input at time $t$ and $\sigma(\cdot)$ represents the sigmoid function.
%while $h()$ represents the hyperbolic tangent.
Eq. (\ref{lstm_eq5}) is frequently found in the literature as $\vec h^{(t)} = \vec c^{(t)} \odot tanh( \vec o^{(t)})$ with $tanh(\cdot)$ applied to the \textit{output} gate. In this work, we follow the image captioning LSTM proposed in \cite{Vinyals_2017} which removes the $tanh(\cdot)$ from the \textit{output} gate and therefore we end up with Eq. (\ref{lstm_eq5}). 
%The model presented in \cite{Vinyals_2017} is used as our primary case study to evaluate the proposed framework in Section \ref{eval_sec}.
%Nevertheless, we follow the form shown in Eq. (\ref{lstm_eq5}) that removes the $tanh(\cdot)$ to stay aligned with \cite{Vinyals_2017}, that is used as a case study in this work.
Finally, all the $\boldsymbol{W}$ matrices denote the weight matrices that contain the trainable parameters of the model, which are assumed to be provided.

%On these weight matrices we will perform SVD in order to investigate how can we approximate the original weight matrix, by taking low-rank approximations. In our computations we concatenate the input with the previous output in one vector, so that we have to deal only with one matrix per gate, on which 4 matrices we will perform SVD.

% * <stylianos.venieris10@imperial.ac.uk> 2017-11-21T10:55:18.501Z:
% 
% The background section on BLEU was good. I rephrased some parts to make it more clear to the reader.
% 
% ^.
% The main idea is that it searches for common words or blocks of words (grams) between a candidate translation and a reference text, since that follows the notion of a good candidate translation. The score is then calculated by taking the ratio of the common n-grams versus all the n-grams of the reference. The usual BLEU metric used is the one that uses up to 4-grams in order to check for similarity.

\vspace{-0.5cm}
\section{Related Work}
\vspace{-0.4cm}

The effectiveness of RNNs % in sequence modelling tasks 
has attracted the attention of the architecture and reconfigurable computing communities. Li et al. \cite{Li_2015} proposed an FPGA-based accelerator for the training of an RNN language model. In \cite{Nurvitadhi_2016}, the authors focus on the optimised deployment of the Gated Recurrent Unit (GRU) model \cite{Chung_2014} in data centres with server-grade FPGAs, ASICs, GPUs and CPUs and propose an algorithmic memoization-based method to reduce the computational load at the expense of higher memory consumption. The authors of \cite{Chang_2017} present an empirical study of the effect of different architectural designs on the computational resource, on-chip memory capacity and off-chip memory bandwidth requirements of an LSTM model. Finally, Guan et al. \cite{Guan_2017} proposed an FPGA-based LSTM accelerator optimised for speech recognition on a Xilinx VC707 FPGA platform.

From an algorithmic perspective, recent works have followed a model-hardware co-design approach. Han et al. \cite{Han_2017} proposed an FPGA-based speech recognition engine that employs a load-balance-aware compression scheme in order to compress the LSTM model size. Wang et al. \cite{Wang_2017} presented a method that addresses compression at several levels including the use of circulant matrices for three of the LSTM gates and the quantisation of the trained parameters, together with the corresponding ASIC-based hardware architecture. Zhang et al. \cite{Zhang_2017} presented an FPGA-based accelerator for a Long-Term Recurrent Convolutional Network (LRCN) for video footage description that consists of a CNN followed by an LSTM. Their design focuses on balancing the resource allocation between the layers of the LRCN and pruning the fully-connected and LSTM layers to minimise the off-chip memory accesses. \cite{Han_2017}, \cite{Wang_2017} and \cite{Zhang_2017} deviate from the faithful LSTM mapping of previous works but also require a retraining step in order to compensate for the introduced error of each proposed method. Finally, He et al. \cite{He_2015} investigated algorithmic strategies for CNN model selection under computation time constraints for both training and testing.

Our work differs from the majority of existing efforts by proposing a hardware architecture together with an approximate computational method for LSTMs that is application-aware and tunable with respect to the required computation time and the application-level error. Our framework follows the same spirit as \cite{Han_2017}\cite{Wang_2017}\cite{Zhang_2017} by proposing an approximation to the model, but in contrast to these methods our scheme does not require a retraining phase for the model and assumes no access to the training set, while compensating for the induced error by means of iterative refinement, making it suitable for applications %where a result needs to be evaluated in a given time budget.
where the dataset is privacy-critical and the quality of the approximation improves as the time availability increases.

% * <stylianos.venieris10@imperial.ac.uk> 2017-11-07T17:47:51.647Z:
% 
% We should focus on two topics: (i) RNNs on FPGAs and (ii) approximations that are similar to ours for neural networks. 
% For (ii) there are a handful of works that use low-rank approximations and we should mention them and position our work with respect to them.
% ADD Jason Cong's paper on LSTMs.
% 
% ^.
% * <a.kouris16@imperial.ac.uk> 2017-11-07T19:57:56.250Z:
% 
% Work on acceleration of LSTMs in general (in GPUs / ASICs) could also be of interest.
% 
% ^.

\vspace{-0.4cm}
\section{Methodology}
\vspace{-0.3cm}

\begin{figure} 
	\centering
	\includegraphics[width=0.8\columnwidth, trim={0 11mm 0 0},clip]{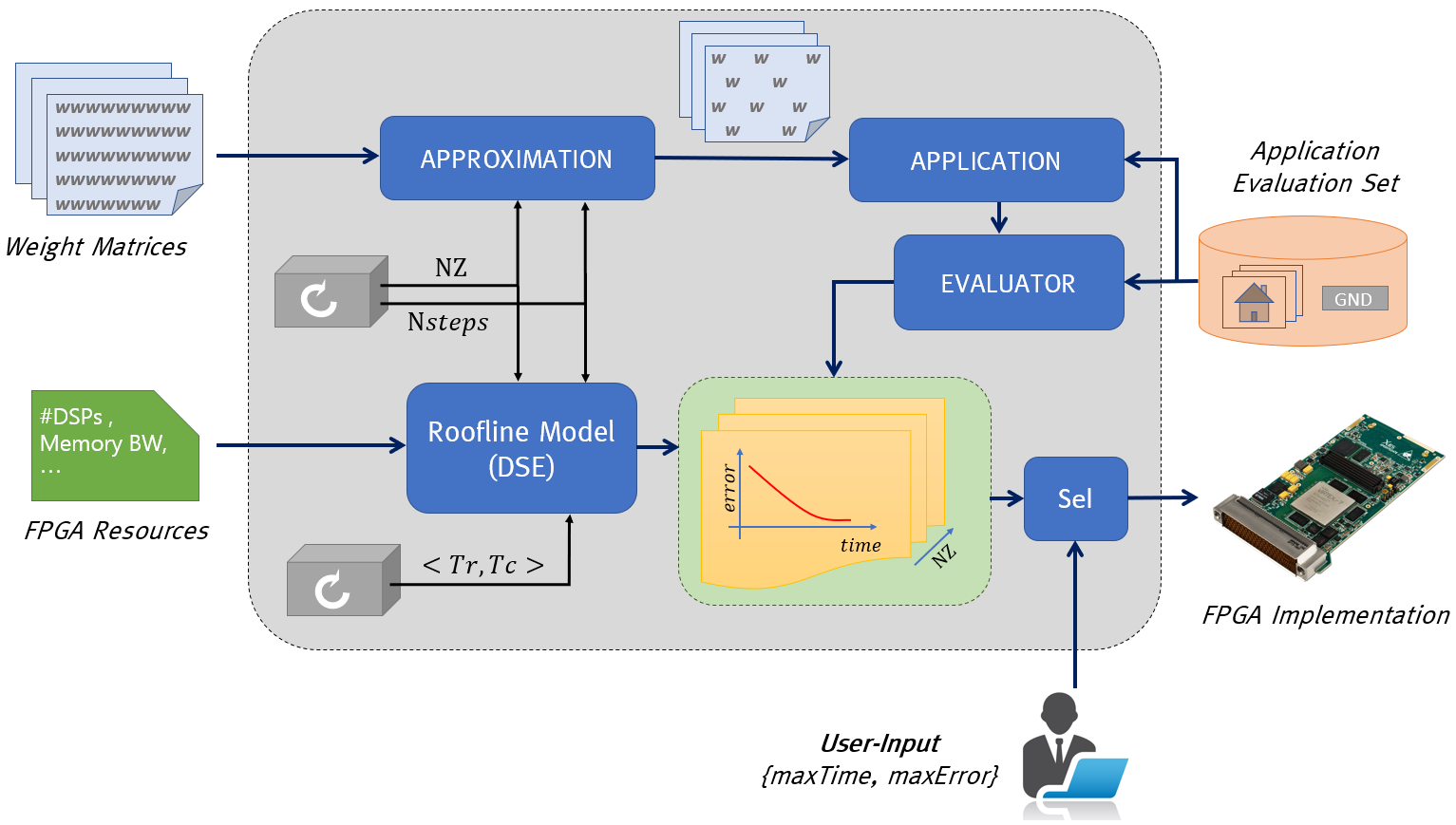}
    \vspace{-0.3cm}
	\caption{Design flow of the proposed framework}
	\label{fig:flow}
    \vspace{-0.9cm}
\end{figure}

In this section, the main components of the proposed framework are presented (Fig. \ref{fig:flow}). %The design flow of our system is illustrated in . 
Given an LSTM model with its set of weight matrices and a small application evaluation set, the proposed system searches for an appropriate approximation scheme that meets the application's needs, by applying low-rank compression and pruning on the model. The design space is traversed by means of a roofline model to determine the highest performing configuration of the proposed architecture on the target FPGA. %Concurrently, performance modelling is applied on each instance of this search on a design space of hardware architectures, to determine the highest performing design point that can be mapped to the given FPGA platform. 
In this manner, the trade-off between computation time and application-level error is explored for different approximation schemes. The design point to be implemented on the device is selected based on user-specified requirements of the maximum computation time or application-level error tolerance.
\vspace{-0.5cm}

% * <a.kouris16@imperial.ac.uk> 2017-11-20T21:51:45.444Z:
% 
% The current version of figures is a low-quality draft preview. 
% 
% ^ <a.kouris16@imperial.ac.uk> 2017-11-23T13:07:44.462Z.

%This will be the main section, which I want to divide in three subsections.

\vspace{-0.3cm}

\subsection{Approximations for LSTMs}
\vspace{-0.2cm}
At the core of an LSTM's computational workload lies the matrix-vector multiplications in each of the four gates. Neural networks have been extensively studied to have redundancy in terms of their trained parameters \cite{Denil_2013}. To reduce the computational demands of the LSTM, we propose an approximate computing scheme that enables the tuning between computational cost and application-level accuracy. The proposed approach exploits the statistical redundancy of the LSTM by acting at two levels: (i) approximating the weight matrices with a low-rank, SVD-based decomposition and (ii) pruning the network by sparsifying the weight matrices based on an importance criterion of their elements.

%a low-rank, SVD-based approximation and (ii) a sparcification method for the weight matrices.

%In this respect, we leverage the redundancy in the LSTM weight matrices of the gates by means of (i) a tunable low-rank, SVD-based approximation method and (ii) a weights sparsification scheme.

\vspace{-0.4cm}

\subsubsection{Low-rank approximation.}
Based on the set of LSTM equations (\ref{lstm_eq1})-(\ref{lstm_eq4}), each gate consists of two weight matrices corresponding to the current input and previous output vectors respectively. In our scheme, we construct an augmented matrix by concatenating the input and output weight matrices as shown in \mbox{Eq. (\ref{aug_matrix_eq})}. Similarly, we concatenate the input and previous output vectors \mbox{(Eq. (\ref{aug_input_eq}))} and thus the overall gate computation is given by Eq. (\ref{aug_gate_eq}).
\vspace{-0.2cm}
\begin{align}
\label{aug_input_eq}
\tilde{\boldsymbol{x}}^{(t)} &= \left[\boldsymbol{x^{(t)}}^T ~~ \boldsymbol{h}^{(t-1)T}\right]^T 
\end{align}
\vspace{-0.2cm}
\begin{align}
\label{aug_matrix_eq}
\boldsymbol{W}_{i} &= \left[\boldsymbol{W}_{ix} ~~ \boldsymbol{W}_{ih}\right], ~~~ \forall i \in [1,4] \\
\label{aug_gate_eq}
\boldsymbol{y}_i &= nonlin(\boldsymbol{W}_i \tilde{\boldsymbol{x}}^{(t)}), ~~ \forall i \in [1,4]
\vspace{-0.9cm}
\end{align}
where $nonlin(\cdot)$ is either the sigmoid function $\sigma(\cdot)$ or $tanh$. In this way, a single weight matrix is formed for each gate, denoted by \mbox{$\boldsymbol{W}_i \in \mathbb{R}^{R \times C}$} for the $i_{th}$ gate. We perform a full SVD decomposition on the four augmented matrices independently as $\boldsymbol{W}_i = \boldsymbol{U}_i \boldsymbol{\Sigma}_i \boldsymbol{V}_i^T, ~\forall i \in [1,4]$, where $\boldsymbol{U}_i \in \mathbb{R}^{R \times R}$, $\boldsymbol{\Sigma}_i \in \mathbb{R}^{R \times C}$ and $\boldsymbol{V}_i \in \mathbb{R}^{C \times C}$ and utilise a rank-1 approximation to obtain $\widetilde{\boldsymbol{W}_i} = \sigma_1^i \boldsymbol{u}_1^i \boldsymbol{v}_1^{iT}$ by keeping the singular vectors that correspond to the largest singular value.

% We perform a full SVD decomposition on the four augmented weight matrices independently and calculate the orthonormal vectors along with their singular values off-line as $\boldsymbol{W}_i = \boldsymbol{U}_i \boldsymbol{\Sigma}_i \boldsymbol{V}_i^T, ~\forall i \in [1,4]$, where $\boldsymbol{U}_i \in \mathbb{R}^{R \times C}$, $\boldsymbol{\Sigma}_i \in \mathbb{R}^{C \times C}$ and $\boldsymbol{V}_i \in \mathbb{R}^{R \times C}$. We utilise a rank-1 approximation by keeping the singular vectors $\boldsymbol{u}_1^i$ and $\boldsymbol{v}_1^i$ that correspond to the largest singular value, $\sigma_1$, and approximate it as $\widetilde{\boldsymbol{W}_i} = \sigma_1^i \boldsymbol{u}_1^i \boldsymbol{v}_1^{iT}$.

%, as given by Eq. (\ref{svd_eq}).
% \begin{eqnarray}
% \label{svd_eq}
% \boldsymbol{W}_i = \boldsymbol{U}_i \boldsymbol{\Sigma}_i \boldsymbol{V}_i^T, ~~~\forall i \in [1,4]
% \end{eqnarray}

\vspace{-0.4cm}

\subsubsection{Pruning by means of network sparsification.}
The second level of approximation on the LSTM comprises the structured pruning of the connectivity between neurons. With each neural connection being captured as an element of the weight matrices, we express network pruning as sparsification applied on the augmented weight matrices (Eq. (\ref{aug_matrix_eq})). To represent a sparse LSTM, we introduce four binary mask matrices $\boldsymbol{F}_i \in \{0,1\}^{R \times C}$, $i \in [1,4]$, with each entry representing whether a connection is pruned or not. Overall, we employ the following notation for a (weight, mask) matrix pair $\left\{\boldsymbol{W}_{i}, \boldsymbol{F}_i ~|~ i \in [1,4] \right\}$.

%TODO Motivation for structured pruning.

In the proposed scheme, we explore sparsity with respect to the connections per output neuron and constrain each output to have the same number of inputs. We cast the LSTM pruning to an optimisation problem of the following form.
\vspace{-0.2cm}
\begin{equation}
\label{pruning_eq}
\min_{\boldsymbol{F}_i} || \boldsymbol{W}_i - \boldsymbol{F}_i \odot \boldsymbol{W}_i ||^2_2,~~~ \text{s.t.} ~~||\boldsymbol{f}_{j}^i||_0 = \text{NZ},~\forall i \in [1,4], \forall j \in [1,n] 
\vspace{-0.4cm}
\end{equation}
where $\boldsymbol{f}_{j}^i$ is the $j_{th}$ row of $\boldsymbol{F}_i$ and NZ is the number of non-zero elements on each row of $\boldsymbol{F}_i$. $||\cdot||_0$ is the $l_0$ pseudo-norm denoting the number of non-zero entries in a vector. The solution to the optimisation problem in Eq. (\ref{pruning_eq}) is given by keeping the NZ elements on each row of $\boldsymbol{W}_i$ with the highest absolute value and setting their indices to 1 in $\boldsymbol{F}_i$.

In contrast to the %majority of 
existing approaches, the proposed pruning method does not employ a retraining step and hence removes the requirement for the training set, which is important for privacy-critical applications, as well as the computationally expensive step of retraining. Even though our sparsification method does not explicitly capture the impact of pruning on the application-level accuracy, our design space exploration, detailed in Section \ref{sec:dse}, searches over different levels of sparsity and as a result it explores the effect of pruning on the application.

\vspace{-0.4cm}

\subsubsection{Hybrid compression and pruning.}
By applying both low-rank approximation and pruning, we end up with the following weight matrix approximation:
\vspace{-0.2cm}
\begin{equation}
\widetilde{\boldsymbol{W}}_i = \boldsymbol{F}_i \odot (\sigma_1^i \boldsymbol{u}_1^i \boldsymbol{v}_1^{iT})
\vspace{-0.1cm}
\end{equation}
In this setting, for the $i_{th}$ gate the ranking of the absolute values in each row of the rank-1 approximation $\sigma_1^i \boldsymbol{u}_1^i \boldsymbol{v}_1^{iT}$ depends only on $\boldsymbol{v}_1^{i}$, with each element of the vector $\sigma_1^i \boldsymbol{u}_1^i$ operating as a scaling factor for all elements of a row. 
Therefore,  for the $i_{th}$ gate all the rows of $\boldsymbol{F}_i$ become identical and hence can be represented by a single mask vector $\boldsymbol{f}^i \in \{0,1\}^{C}$. This leads to a weight matrix with zeros along $n$$-$$\text{NZ}$ of its columns, which is described by the following expression: % Computationally, this translates to the following equivalent expression:
\vspace{-0.2cm}
\begin{align}
\label{hybrid_vec_form_eq}
\widetilde{\boldsymbol{W}}_i &= \sigma_1^i \boldsymbol{u}_1^i (\boldsymbol{f}^i \odot \boldsymbol{v}_1^{iT}) \\
\label{gate_comp_eq}
\tilde{\boldsymbol{y}}_i &= \sum_{n=1}^{N_{steps}} \left\{ \sigma_1^{i(n)} \boldsymbol{u}_1^{i(n)} \left( (\boldsymbol{f}^{i(n)} \odot \boldsymbol{v}_1^{i(n)})^T \tilde{\boldsymbol{x}}^{(t)} \right) \right\}
\vspace{-1.6cm}
\end{align}

In order to obtain a refinement mechanism, we propose an iterative algorithm, presented in Algorithm \ref{iter_refine_alg}, that employs both the low-rank approximation and pruning methods to progressively update the weight matrix. On lines 4-6 the first approximation of the weight matrix is constructed by obtaining the rank-1 approximation of the original matrix and applying pruning in order to have NZ non-zero elements on each row, as in Eq. (\ref{hybrid_vec_form_eq}). Next, the weight matrix is refined for $N_{steps}$ iterations, by computing the error matrix $\boldsymbol{E}$ (line 10) and employing its pruned, rank-1 approximation as our update (line 15). 

\vspace{-0.5cm}
\begin{algorithm}[H]
	\footnotesize
	\caption{Iterative LSTM Model Approximation}
	\textbf{Inputs:}
	\begin{algorithmic}[1]
		\STATE Weight matrices $\boldsymbol{W}_i \in \mathbb{R}^{R \times C},~\forall i \in [1,4]$ \\
%		\STATE Mask matrices $\boldsymbol{F}_i \in \{0,1\}^{m \times n},~\forall i \in [1,4]$
		\STATE Number of non-zero elements, NZ
		\STATE Number of refinement steps, $N_{steps}$
	\end{algorithmic}
	
	\textbf{Steps:}
		\begin{algorithmic}[1]
		\STATE - - For all gates - -
		\FOR{$i$ = 1 to 4}
			\STATE - - Initialise weight matrix approximation - -
			\STATE $\left[\boldsymbol{u}_1^{i(0)}, \sigma_1^{i(0)}, \boldsymbol{v}_1^{i(0)}\right] = \text{SVD}(\boldsymbol{W}_i)_1$
			\STATE $\boldsymbol{f}^{i(0)} \leftarrow $ solution to Eq. (\ref{pruning_eq}) for vector $\boldsymbol{v}_1^{i(0)}$
			\STATE $\widetilde{\boldsymbol{W}}_i^{(0)} = \sigma_1^{i(0)} \boldsymbol{u}_1^{i(0)} \left(\boldsymbol{f}^{i(0)} \odot \boldsymbol{v}_1^{i(0)} \right)^T$
%			\STATE $\widetilde{\boldsymbol{W}}_i^{(0)} = \boldsymbol{F}_i \odot \text{SVD}(\boldsymbol{W}_i)_1$
			\STATE - - Apply refinements - -
			\FOR{$n$ = 1 to $N_{steps}$}
				\STATE - - Compute error matrix - -
				\STATE $\boldsymbol{E} = \boldsymbol{W}_i - \widetilde{\boldsymbol{W}}_i^{(0)}$
				\STATE - - Compute refinement - -
				\STATE $\left[\boldsymbol{u}_1^{i(n)}, \sigma_1^{i(n)}, \boldsymbol{v}_1^{i(n)}\right] = \text{SVD}(\boldsymbol{E})_1$
				\STATE $\boldsymbol{f}^{i(n)} \leftarrow $ solution to optimisation problem (\ref{pruning_eq}) for vector $\boldsymbol{v}_1^{i(n)}$
				\STATE - - Update weight matrix approximation - -
				\STATE $\widetilde{\boldsymbol{W}}_i^{(n)} = \widetilde{\boldsymbol{W}}_i^{(n-1)} + \sigma_1^{i(n)} \boldsymbol{u}_1^{i(n)} \left(\boldsymbol{f}^{i(n)} \odot \boldsymbol{v}_1^{i(n)} \right)^T$
%				\STATE $\widetilde{\boldsymbol{W}}_i^{(t)} = \widetilde{\boldsymbol{W}}_i^{(t-1)} + \boldsymbol{F}_i \odot \boldsymbol{R}$
%				\STATE $\widetilde{\boldsymbol{W}}_i^{(t)} = \widetilde{\boldsymbol{W}}_i^{(t-1)} + \boldsymbol{F}_i \odot \text{SVD}(\widetilde{\boldsymbol{W}}_i^{(t-1)})_1$
			\ENDFOR
		\ENDFOR
		\end{algorithmic}
		Notes: (i) SVD$(\boldsymbol{X})_1$ returns the rank-$1$ SVD-based approximation of $\boldsymbol{X}$.
%		\\
%		$~~~~~~~~~~$(ii) $\boldsymbol{F}_i \odot \text{SVD}(\boldsymbol{X})$ can be computed as in Eq. (\ref{hybrid_vec_form_eq}).
		\label{iter_refine_alg}
\end{algorithm}
\vspace{-0.7cm}
Different combinations of levels of sparsity and refinement iterations correspond to different design points in the computation-accuracy space. In this respect, the number of non-zero elements in each binary mask vector and the number of iterations are exposed to the design space exploration as tunable parameters (NZ, $N_{steps}$) to explore the LSTM computation-accuracy trade-off.
\vspace{-0.4cm}

\subsection{Architecture}
\vspace{-0.2cm}

\begin{figure} 
	\vspace{-0.7cm}
 	\includegraphics[width=0.95\columnwidth]{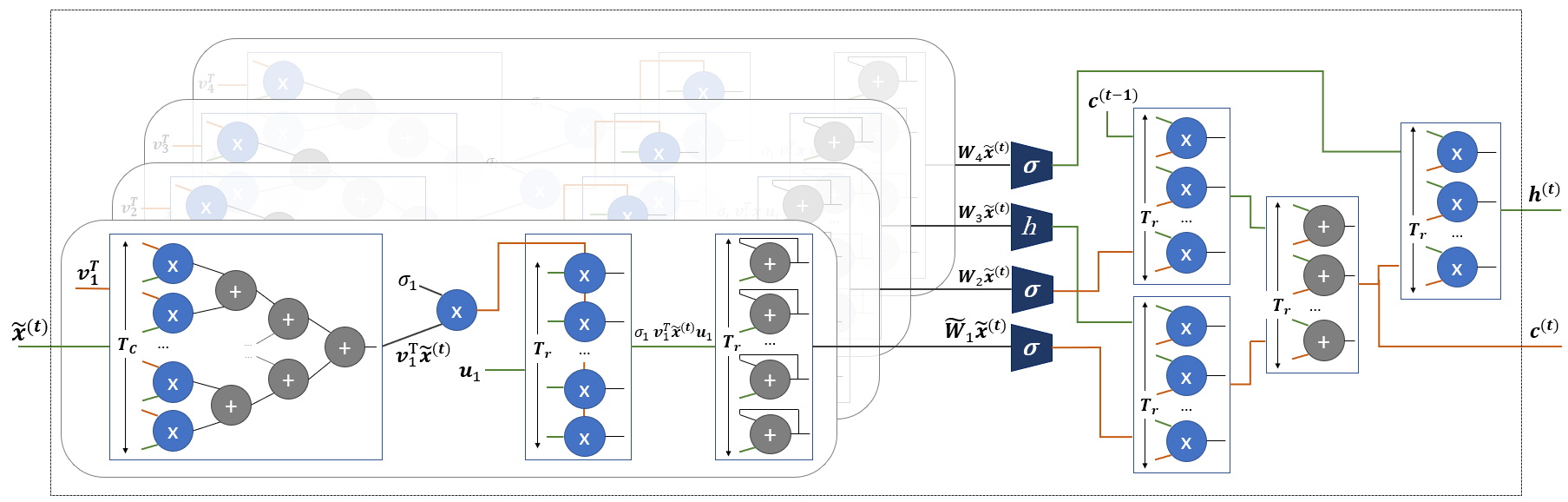}
    \vspace{-0.25cm}
	\caption{Diagram of proposed hardware architecture}
	\label{fig:architecture}
    \vspace{-1.1cm}
\end{figure}

The proposed FPGA architecture for LSTMs is illustrated in Fig. \ref{fig:architecture}. The main strategy of the architecture includes the exploitation of the coarse-grained parallelism between the four LSTM gates and is parametrised with respect to the fine-grained parallelism in the dot product and elementwise operations of the LSTM, allowing for a compile-time tunable performance-resource trade-off.

\vspace{-0.45cm}

\subsubsection{SVD and Binary Masks Precomputation.}
In Algorithm \ref{iter_refine_alg}, the number of refinement iterations ($N_{steps}$), the level of sparsity (NZ) and the trained weight matrices %the trained model, i.e. the weight matrices of each gate ($\boldsymbol{W}_i$),
are data-independent and known at compile time. As such, the required SVD decompositions along with the corresponding binary masks are precomputed for all $N_{steps}$ iterations at compile time. As a result, the singular values $\sigma_1^{i(n)}$, the vectors $\boldsymbol{u}_1^{i(n)}$ and only the non-zero elements of the sparse $\boldsymbol{f}^{i(n)} \odot \boldsymbol{v}_1^{i(n)}$ are stored in the off-chip memory, so that they can be looked-up at run time.

\vspace{-0.2cm}

\subsubsection{Inter-gate and Intra-gate Parallelism.}
In the proposed architecture, each gate is allocated a dedicated \textit{hardware gate unit} with all gates operating in parallel. At each LSTM time-step $t$, a hardware gate unit is %responsible for
computing its output by performing $N_{steps}$ refinement iterations as in Eq. (\ref{gate_comp_eq}). At the beginning of the time-step, the current vector $\tilde{\boldsymbol{x}}^{(t)}$ is stored on-chip as it will be reused in each iteration by all four gates. The vectors $\boldsymbol{u}_1^{i(n)}$ and $\boldsymbol{v}_1^{i(n)}$ for each gate, along with their singular values $\sigma_1^{i(n)}$, are streamed in the architecture from the off-chip memory in a tiled manner. $\boldsymbol{u}_1^{i(n)}$ and $\boldsymbol{v}_1^{i(n)}$ are tiled with tile sizes of $T_r$ and $T_c$ respectively, leading to $\frac{R}{T_r}$ and $\frac{C}{T_c}$ tiles sequentially streamed in the architecture. %Vectors $\boldsymbol{u}_1^{i(n)}$ and $\boldsymbol{v}_1^{i(n)}$ are tiled with tile sizes of $T_r$ and $T_c$ respectively. In this manner, $\frac{R}{T_r}$ and $\frac{C}{T_c}$ tiles are sequentially streamed in the architecture respectively. 

At each gate, a dot product unit is responsible for computing the dot product of the current tile of $\boldsymbol{v}_1^{i(n)}$ with the corresponding elements of the input $\tilde{\boldsymbol{x}}^{(t)}$. The dot product unit is unrolled by a factor of $T_c$ in order to process one tile of $\boldsymbol{v}_1^{i(n)}$ per cycle. After accumulating the partial results of all the $\frac{C}{T_c}$ tiles, the result is produced and multiplied with the scalar $\sigma_1^{i(n)}$. The multiplication result is passed as a constant operand in a multiplier array, with $\boldsymbol{u}_1^{i(n)}$ as the other operand. The multiplier array has a size of $T_r$ in order to match the tiling of $\boldsymbol{u}_1^{i(n)}$. As a final stage, an array of $T_r$ accumulators performs the summation across the $N_{steps}$ iterations, as expressed in Eq. (\ref{gate_comp_eq}) to produce the final gate output.

The outputs from the \textit{input}, \textit{forget} and \textit{output} gates are passed through a sigmoid unit while the output of the \textit{cell} gate is passed through a $tanh$ unit. After the nonlinearities stage, the produced outputs are multiplied element-by-element as dictated by the LSTM equations to produce the cell state $\boldsymbol{c}^{(t)}$ \mbox{(Eq. (\ref{lstm_eq4}))} and the current output vector $\boldsymbol{h}^{(t)}$ (Eq. (\ref{lstm_eq5})). The three multiplier arrays and the one adder array all have a size of $T_r$ to match the tile size of the incoming vectors and exploit the available parallelism.

\vspace{-0.35cm}

\section{Design Space Exploration}
\label{sec:dse}

\vspace{-0.25cm}

Having parametrised the proposed approximation method over NZ and $N_{steps}$ and its underlying architecture over NZ and tile sizes ($T_r$, $T_c$), corresponding metrics need to be employed for exploring the effects of each parameter on performance and accuracy. The approximation method parameters are studied based on an application-level evaluation metric (discussed in Section \ref{sec:dse:application}), that measures the impact of each applied approximation on the accuracy of the target application. In terms of the hardware architecture, roofline performance modelling is employed for exhaustively exploring the design space formed by all possible tile size combinations, to obtain the highest performing design point (discussed in Section \ref{sec:dse:roofline}). Based on those two metrics, the computation time-accuracy trade-off is explored. 

\vspace{-0.45cm}

\subsection{Roofline Model}
\label{sec:dse:roofline}
\vspace{-0.15cm}

  The design space of architectural configurations for all tile size combinations of $T_r$ and $T_c$ is explored exhaustively by performance modelling. The roofline model \cite{williams2009roofline} is used to develop a performance model for the proposed architecture by relating the peak attainable performance (in terms of throughput) for each configuration on a particular FPGA device with its operational intensity, which relates the ratio of computational load to off-chip memory traffic. Based on this model, each design point's performance can be bounded either by the peak platform throughput or by the maximum performance that the platform's memory system can support. %The proposed architecture is implemented by selecting the tile sizes ($T_r$, $T_c$) that correspond to the highest performing design point that is supported by the target platform.
In this context, roofline models are developed for predicting the maximum attainable performance for varying levels of pruning (NZ). %a varying number of non-zero elements in each binary mask (NZ). 

Given a tile size pair, the performance of the architecture is calculated as:
\begin{equation}
\label{eq:Perf}
\begin{split}
	Perf (ops/s) = \frac{workload (ops/input)}{II (cycles/input)}  
   = \frac{4 N_{steps} (2NZ+2R+1) + 37R }{max(N_{ steps} max( \frac{R}{T_r} , \frac{NZ}{T_c} ) , 37 \frac{R}{T_r}  )}
 \end{split}
\end{equation}
%\begin{equation}
%\begin{split}
%	Performance (ops/s) = \frac{workload (ops/input)}{II (cycles/input)}  \\ \\
 %  = \frac{4 N_{steps} (2C+2R+1) + 37R }{max(N_{ steps} max( \frac{R}{T_r} , \frac{C}{T_c} ) , 37 \frac{R}{T_r}  )}
 %\end{split}
%\end{equation}
where each gate performs $2NZ+2R+1$  operations per iteration and 37$R$ accounts for the rest of the operations to produce the final outputs. The initiation interval ($II$) is determined based on the slowest between the gate stage and the rest of the computations. Similarly, a gate's initiation interval depends on the slowest between the dot product unit and the multiplier array (Fig. \ref{fig:architecture}).

Respectively, the operational intensity of the architecture, also referred to in the literature as Computation-to-Communication ratio (CTC), is formulated as:
\begin{equation}
\label{eq:ctc}
\begin{split}
	CTC(ops/byte) = \frac{operations (ops)}{mem\ access (bytes)} 
   = \frac{ 4 N_{steps} (2NZ + 2R + 1) + 37R }{4 (4 N_{steps} (NZ + R + 1) + 2R)}
 \end{split}
\end{equation}
%\begin{equation}
%\begin{split}
%	OpIntensity (ops/byte) = \frac{num \ of \ operations (ops)}{amount \ of \ memory \ access (bytes)}  \\ \\
  % = \frac{ 4 N_{steps} (2C + 2R + 1) + 37R }{4 (4 N_{steps} (C + R + 1) + 2R)}
 %\end{split}
%\end{equation}
where the memory transfers include the singular vectors and the singular value for each iteration of each gate and the write-back of the output and the cell state vectors to the off-chip memory. The augmented input vector $\tilde{\boldsymbol{x}}^{(t)}$ is stored on-chip in order to be reused across the $N_{steps}$ iterations. All data are represented with a single-precision floating-point format and require four bytes.

%Finally, the on-chip memory requirements of the architecture can be described as:
%\begin{equation}
% \begin{split}
%	 on\ chip(bytes) = 4 \left(4 (R+NZ+1) +2R \right)
%  \end{split}
%\end{equation}
%where at each instant the singular vectors and the singular value of each gate that correspond to the current iteration should be fetched to the on-chip memory, as well as the computed output and cell state vectors which are kept on-chip until the end of the iteration.
%For each number or non-zero elements NZ, the design space consists of $|T_c||T_r|$ design points. By definition $T_r$ is bounded by the number of rows in the weight matrices ($T_r \leq R$), while $T_c$ is bounded by the number of non-zero elements of the approximation ($T_c \leq $NZ). Thus for each value of NZ the total number of design points can be calculated as: 
%\begin{equation}
%	|T_c||T_r| = \sum_{i=1}^{NZ} i R = R \frac{1+NZ}{2} NZ 
%\end{equation}
%Providing that the maximum number of non-zero elements examined is bounded by the number of columns of the weight matrices ($NZ\leq C$), the total number of design points across all possible values of NZ examined are $|NZ||T_c||T_r|$ which can be also calculated analytically as:
%\begin{equation}
%	|NZ||T_c||T_r| = \sum_{j=1}^{C} R \frac{1+j}{2} j  = R \frac{1+\frac{1+C}{2}C}{2}C = R\frac{C^3+3C^2}{4}
%\end{equation}
The number of  design points allows enumerating all possible tile size combinations for each number of non-zero elements and obtaining the performance and CTC values for the complete design space. Based on the target platform's peak performance, memory bandwidth and on-chip memory capacity, the subspace containing the platform-supported design points can be determined. The proposed architecture is implemented by selecting the tile sizes ($T_r$, $T_c$) that correspond to the highest performing design point within that subspace. 
%within which the design that would achieve maximum throughput is selected. 

\vspace{-0.3cm}

\subsection{Evaluating the Impact of Approximations on the Application}
\label{sec:dse:application}
\vspace{-0.2cm}

The proposed framework requires a metric that would allow us to measure the impact of the applied approximations on the application-level accuracy for different (NZ, $N_{steps}$) pairs. In our methodology, the error induced by our approximation methods is measured by running the target application end-to-end over an evaluation set with both our approximated weight matrices given a selected (NZ, $N_{steps}$) pair and with the %reference model.
original pretrained LSTM, acting as a reference model. By treating the output of the reference model as the ground truth, an application-specific metric is employed that assesses the quality of the output that was generated by the approximate model, exploring in this way the relationship between the level of approximation and the application-level accuracy.

\vspace{-0.4cm}

\section{Evaluation}
\label{eval_sec}

\vspace{-0.3cm}

%To evaluate the proposed framework, we focus on the image captioning system presented by Vinyals et al. \cite{Vinyals_2017} which won the 2015 MSCOCO image captioning challenge. The system employs a convolutional neural network in order to encode input images and feeds the image encodings to an LSTM model to produce captions. In the proposed LSTM, each gate consists of two $512 \times 512 $ weight matrices, leading to a ($512 \times 1024$) augmented weight matrix per gate with $R=512$ and $C=1024$. For the design space exploration of our system, the evaluation set is obtained from the validation set of the Common Objects in Context (COCO) dataset\footnote{http://cocodataset.org}.
The image captioning system presented by Vinyals et al. \cite{Vinyals_2017} (winner of the 2015 MSCOCO challenge) is examined as a case study for evaluating the proposed framework. Input images are encoded by a CNN and fed to a trained LSTM model to predict corresponding captions. In the proposed LSTM, each gate consists of two $R \times R$ weight matrices, leading to a ($R \times C$) augmented weight matrix per gate with $R=512$  and $C=2R$ for a total of 2.1 M parameters. To determine the most suitable approximation scheme, we use a subset of the validation set of the Common Objects in Context (COCO) dataset\footnote{http://cocodataset.org}, consisting of 35 images. %as the evaluator's dataset. 
To obtain image captions that will act as ground truth for the evaluation of the proposed approximation method, the reference image captioning application is executed end-to-end over the evaluation set, using TensorFlow\footnote{https://www.tensorflow.org}. As a metric of the effect of low-rank approximation and pruning applied on the LSTM model, we select Bi-lingual Evaluation Understudy (BLEU) \cite{Papineni_2001}, which is commonly employed for the  evaluation of machine translation's quality by measuring the number of matching words, or ``blocks of words", between a reference and a candidate translation. Due to space limitations, more information about adopting BLEU as a quality metric for image captioning can be found in \cite{Vinyals_2017}.

\vspace{-0.1cm}

\subsubsection{Experimental Setup.}
\vspace{-0.3cm}
In our experiments, we target the Xilinx Zynq ZC706 board. All hardware designs were synthesised and placed-and-routed with Xilinx Vivado HLS and Vivado Design Suite (v17.1) with a clock frequency of \mbox{100 MHz.} Single-precision floating-point representation was used in order to comply with the typical precision requirements of LSTMs as used by the deep learning community. Existing work \cite{Li_2015}\cite{Han_2017} has studied precision optimisation in specific LSTM applications, which constitutes a complementary method to our framework as an additional tunable parameter for the performance-accuracy trade-off.

%We developed our hardware design on the Xilinx ZC706 board, using the Vivado HLS tool, version 2017.1.

% \subsubsection{Benchmark Application.}
% As a case study, the LSTM-based image captioning system presented by Vinyals et al. \cite{Vinyals_2017} is targeted. Their system employs a convolutional neural network in order to encode input images and feeds the image encodings to an LSTM model to produce captions. In the proposed LSTM, each gate consists of two $512$$\times$$512$ weight matrices, leading to a ($512$$\times$$1024$) augmented weight matrix per gate with $R$$=$$512$ and $C$$=$$1024$. The target dataset is the Common Objects in Context dataset by Microsoft\footnote{http://cocodataset.org}

% Our target application is the image captioning application from the Show and Tell paper by Google <Insert reference here>, which combines one CNN for decoding images, and one LSTM cell for producing captions. The LSTM is fed initially with the output of the CNN, embedded into the input space of the LSTM, and then the inputs for the next time steps are words from the vocabulary that are also embedded into the input space of the LSTM. The best captions are determined through beam search of the outputs of the LSTM. The LSTM itself has two weight matrices for each of the 4 gates, with dimensions 512x512, since 512 are the hidden units of the LSTM for this application.

\vspace{-0.4cm}

\subsubsection{Baseline Architecture.}
\begin{figure} 
	\vspace{-0.65cm}
	\centering
 	\includegraphics[width=0.99\columnwidth, trim={11mm 0mm 11mm 0},clip]{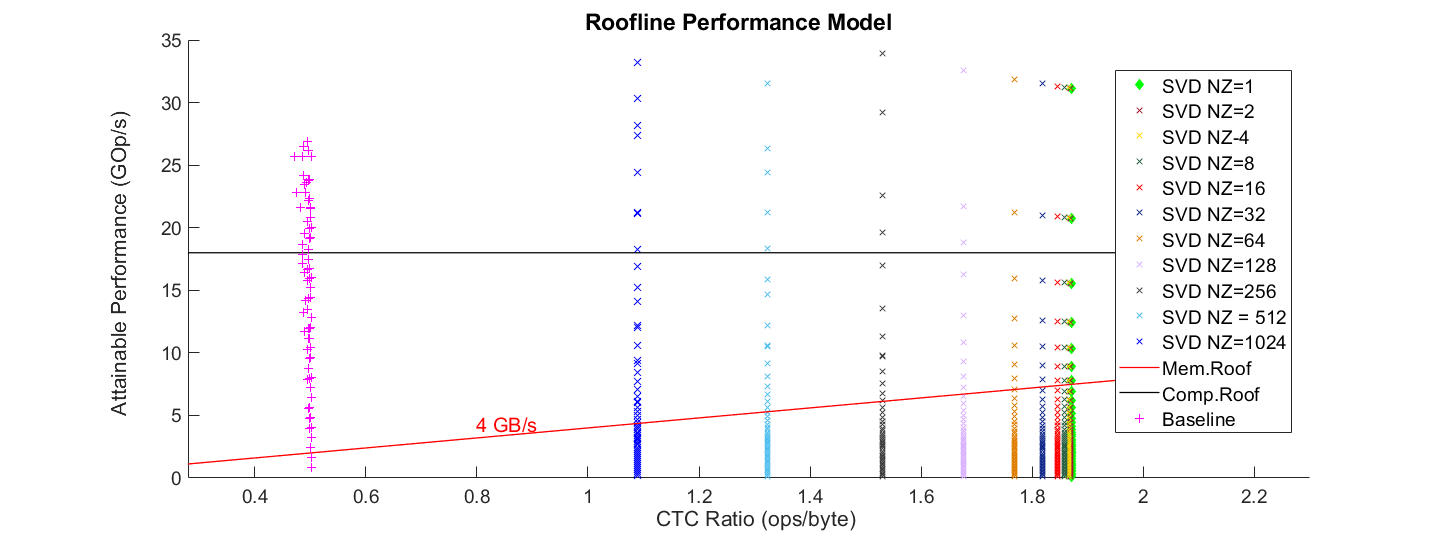}
    \vspace{-0.3cm}
	\caption{Roofline model of the proposed and baseline architectures on the ZC706 board}
	\label{fig:roofline}
    \vspace{-1.1cm}
\end{figure}

%As a reference hardware design that would not employ our approximation scheme, a baseline architecture is designed and implemented. The baseline architecture follows a faithful implementation of the LSTM model without approximations and consists of four parallel hardware gate units that perform matrix-vector multiplication in a tiled manner. The baseline design is parametrised with respect to the tiling along the rows ($T_r)$ and columns ($T_c$) of the weight matrices and the roofline model is used to obtain the highest performing configuration ($T_r$, $T_c$), which is (2, 1) for the target platform and consumes 308 DSPs (34\%), 69 kLUTs (31\%), 437 kFFs (21\%) and \mbox{1,090 18Kbit BRAMs (2\%).} (Fig. \ref{fig:roofline}). To obtain the application-level accuracy of the baseline design under time constrained scenarios, the computation of the LSTM's output is stopped based on factor $T_r$ and the resulting BLEU at each time instant is measured (Fig. \ref{fig:bleu}).
A hardware architecture of a faithful implementation of the LSTM model is implemented to act as a baseline for the proposed system's evaluation. This baseline architecture consists of four gate units, implemented on parallel hardware, that perform matrix-vector multiplication in a tiled manner. Parametrisation with respect to the tiling along the rows ($T_r)$ and columns ($T_c$) of the weight matrices is applied to this architecture and roofline modelling is used to obtain the highest performing configuration ($T_r$, $T_c$), similarly to the proposed system's architecture (Fig. \ref{fig:roofline}). The maximum platform-supported attainable performance was obtained for $T_r=2$ and $T_c = 1$, utilising 308 DSPs (34\%), 69 kLUTs (31\%), 437 kFFs (21\%) and \mbox{26 18kbit BRAMs (2\%)}. As \mbox{Fig. \ref{fig:roofline}} demonstrates, the designs are mainly memory bounded and as a result not all the FPGA resources are utilised. To obtain the application-level accuracy of the baseline design under time constrained scenarios, the BLEU of the intermediate LSTM's output at each tile step of $T_r$ is examined (Fig. \ref{fig:bleu}).

\subsection{Comparisons at Constrained Computation Time}
\begin{figure}
	\vspace{-0.75cm}
	\centering
 	\includegraphics[width=0.97\columnwidth, trim={0 0 0 0},clip]{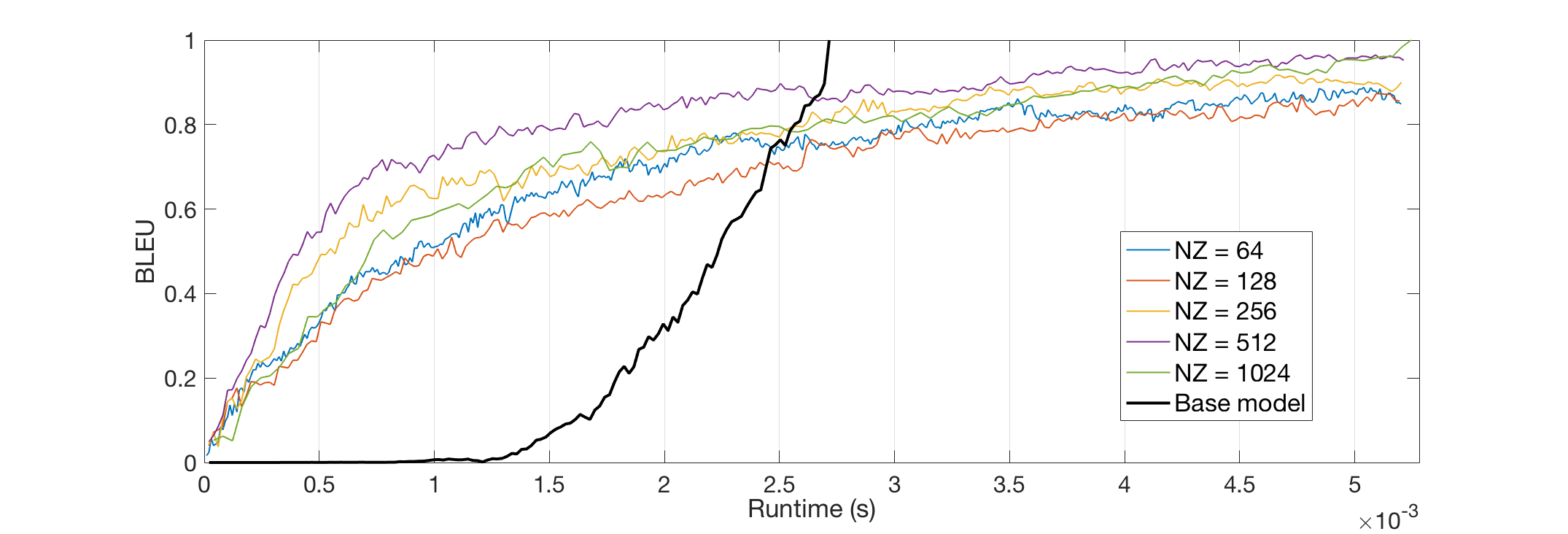}
    \vspace{-0.4cm}
	\caption{BLEU scores over time for all methods}
	\label{fig:bleu}
        \vspace{-0.8cm}
\end{figure}

\vspace{-0.2cm}
This section presents the gains of using the proposed methodology compared to the baseline design under computation time constraints. This is investigated by exploring the design space, defined by (NZ, $T_r$, $T_c$), in terms of (i) performance (Fig. \ref{fig:roofline}) and (ii) the relationship between accuracy and computation time \mbox{(Fig. \ref{fig:bleu}).} As shown in Fig. \ref{fig:roofline}, as the level of pruning increases and NZ becomes smaller, the computational and memory load per refinement iteration becomes smaller and the elementwise operations gradually dominate the computational intensity (Eq. (\ref{eq:ctc})), with the corresponding designs moving to the right of the roofline graph. With respect to the architectural parameters, as the tiling parameters $T_r$ and $T_c$ increase, the hardware design becomes increasingly unrolled and moves towards the top of the roofline graph. In all cases, the proposed architecture demonstrates a higher performance compared to the baseline design reaching up to 3.72$\times$ for a single non-zero element with an average of 3.35$\times$ (3.31$\times$ geo. mean) across the sparsity levels shown in Fig. \ref{fig:roofline}.

To evaluate our methodology in time-constrained scenarios, for each sparsity level the highest performing design of the roofline model is implemented. \mbox{Fig. \ref{fig:bleu}} shows the achieved BLEU score of each design over the evaluation set with respect to runtime, where higher runtime translates to higher number of refinements. In this context, for the target application the design with 512 non-zero elements (50\% sparsity) achieves the best trade-off between performance per refinement iteration and additional information obtained at each iteration. The highest performing architecture with NZ of 512 has a tiling pair of (32, 1) and the implemented design consumes 862 DSPs (95\%), 209 kLUTs (95\%), \mbox{437 kFFs (40\%)} and 34 18kbit BRAMs (3\%). In the BLEU range between 0.4 and 0.8, our proposed system reaches the corresponding BLEU decile up to 6.51$\times$ faster with an average speedup of 4.19$\times$ (3.78$\times$ geo. mean) across the deciles. 

As demonstrated in Fig. \ref{fig:bleu}, the highest performing design of the proposed method (NZ=512) constantly outperforms the baseline architecture in terms of BLEU score at every time instant up to 2.7 ms, at which a maximum BLEU value of 0.9 has been achieved by both methods. As a result, given a specific time budget below 2.7 ms, the proposed architecture achieves a 24.88$\times$ higher BLEU score (geo. mean) compared to the baseline. Moreover, the proposed method demonstrates significantly higher application accuracy during the first 1.5 ms of the computation, reaching up to 31232$\times$ higher BLEU. In this respect, our framework treats a BLEU of 0.9 and a time budget of 2.7 ms as switching points to select between the baseline and the architecture that employs the proposed approximation method and deploys the highest performing design for each case.

\vspace{-0.4cm}

\section{Conclusion}

\vspace{-0.35cm}

The high-performance deployment of LSTMs under stringent computation time constraints poses a challenge in several emerging applications. This paper presents a framework for mapping LSTMs on FPGAs in such scenarios. The proposed methodology applies an iterative approximate computing scheme in order to compress and prune the target network and explores the computation time-accuracy trade-off. A novel FPGA architecture is proposed that is tailored to the degree of approximation and optimised for the target device. This formulation enables the co-optimisation of the LSTM approximation and the architecture in order to satisfy the application-level computation time constraints. Potential future work would include the precision optimisation of the LSTM in order to explore its complementary effect on the performance-accuracy trade-off. 
\vspace{-0.35cm}

\section{Acknowledgements}
\vspace{-0.3cm}
The support of the EPSRC Centre for Doctoral Training in High Performance Embedded and Distributed Systems  (HiPEDS, Grant Reference EP/L016796/1) is gratefully acknowledged. This work is also supported by EPSRC grant 1507723.

\vspace{-0.35cm}

\bibliographystyle{IEEEtran}
\bibliography{ARC2018}

\begin{thebibliography}{10}
\providecommand{\url}[1]{#1}
\csname url@samestyle\endcsname
\providecommand{\newblock}{\relax}
\providecommand{\bibinfo}[2]{#2}
\providecommand{\BIBentrySTDinterwordspacing}{\spaceskip=0pt\relax}
\providecommand{\BIBentryALTinterwordstretchfactor}{4}
\providecommand{\BIBentryALTinterwordspacing}{\spaceskip=\fontdimen2\font plus
\BIBentryALTinterwordstretchfactor\fontdimen3\font minus
  \fontdimen4\font\relax}
\providecommand{\BIBforeignlanguage}[2]{{%
\expandafter\ifx\csname l@#1\endcsname\relax
\typeout{** WARNING: IEEEtran.bst: No hyphenation pattern has been}%
\typeout{** loaded for the language `#1'. Using the pattern for}%
\typeout{** the default language instead.}%
\else
\language=\csname l@#1\endcsname
\fi
#2}}
\providecommand{\BIBdecl}{\relax}
\BIBdecl

\bibitem{byeon2015scene}
W.~Byeon, T.~M. Breuel, F.~Raue, and M.~Liwicki, ``{Scene Labeling with LSTM
  Recurrent Neural Networks},'' in \emph{CVPR}, 2015, pp. 3547--3555.

\bibitem{gregor15}
K.~Gregor, I.~Danihelka, A.~Graves, D.~Rezende, and D.~Wierstra, ``{DRAW: A
  Recurrent Neural Network For Image Generation},'' in \emph{ICML}, 2015, pp.
  1462--1471.

\bibitem{Alahi_2016_CVPR}
A.~Alahi, K.~Goel, V.~Ramanathan, A.~Robicquet, L.~Fei-Fei, and S.~Savarese,
  ``{Social LSTM: Human Trajectory Prediction in Crowded Spaces},'' in
  \emph{CVPR}, 2016.

\bibitem{otte2016recurrent}
S.~Otte \emph{et~al.}, ``{Recurrent Neural Networks for Fast and Robust
  Vibration-based Ground Classification on Mobile Robots},'' in \emph{ICRA},
  2016, pp. 5603--5608.

\bibitem{Vinyals_2017}
O.~Vinyals, A.~Toshev, S.~Bengio, and D.~Erhan, ``{Show and Tell: Lessons
  Learned from the 2015 MSCOCO Image Captioning Challenge},'' \emph{TPAMI}, pp.
  652--663, 2017.

\bibitem{Donahue_2017}
J.~Donahue \emph{et~al.}, ``{Long-Term Recurrent Convolutional Networks for
  Visual Recognition and Description},'' \emph{TPAMI}, vol.~39, no.~4, pp.
  677--691, 2017.

\bibitem{Li_2015}
S.~Li, C.~Wu, H.~Li, B.~Li, Y.~Wang, and Q.~Qiu, ``{FPGA Acceleration of
  Recurrent Neural Network Based Language Model},'' in \emph{FCCM}, 2015, pp.
  111--118.

\bibitem{Nurvitadhi_2016}
E.~Nurvitadhi \emph{et~al.}, ``{Accelerating Recurrent Neural Networks in
  Analytics Servers: Comparison of FPGA, CPU, GPU, and ASIC},'' in \emph{FPL},
  2016, pp. 1--4.

\bibitem{Chung_2014}
J.~Chung \emph{et~al.}, ``{Empirical Evaluation of Gated Recurrent Neural
  Networks on Sequence Modeling},'' in \emph{NIPS Workshop on Deep Learning},
  2014.

\bibitem{Chang_2017}
A.~X.~M. Chang and E.~Culurciello, ``{Hardware Accelerators for Recurrent
  Neural Networks on FPGA},'' in \emph{ISCAS}, 2017, pp. 1--4.

\bibitem{Guan_2017}
Y.~Guan, Z.~Yuan, G.~Sun, and J.~Cong, ``{FPGA-based Accelerator for Long
  Short-Term Memory Recurrent Neural Networks},'' in \emph{ASP-DAC}, 2017, pp.
  629--634.

\bibitem{Han_2017}
S.~Han \emph{et~al.}, ``{ESE: Efficient Speech Recognition Engine with Sparse
  LSTM on FPGA},'' in \emph{FPGA}, 2017, pp. 75--84.

\bibitem{Wang_2017}
Z.~Wang, J.~Lin, and Z.~Wang, ``{Accelerating Recurrent Neural Networks: A
  Memory-Efficient Approach},'' \emph{TVLSI}, vol.~25, no.~10, pp. 2763--2775,
  oct 2017.

\bibitem{Zhang_2017}
X.~Zhang \emph{et~al.}, ``{High-Performance Video Content Recognition with
  Long-Term Recurrent Convolutional Network for FPGA},'' in \emph{FPL}, 2017,
  pp. 1--4.

\bibitem{He_2015}
K.~He and J.~Sun, ``{Convolutional Neural Networks at Constrained Time Cost},''
  in \emph{CVPR}, 2015.

\bibitem{Denil_2013}
M.~Denil, B.~Shakibi, L.~Dinh, M.~A. Ranzato, and N.~de~Freitas, ``{Predicting
  Parameters in Deep Learning},'' in \emph{NIPS}, 2013, pp. 2148--2156.

\bibitem{williams2009roofline}
S.~Williams \emph{et~al.}, ``{Roofline: An Insightful Visual Performance Model
  for Multicore Architectures},'' \emph{Communications of the ACM}, vol.~52,
  no.~4, pp. 65--76, 2009.

\bibitem{Papineni_2001}
K.~Papineni, S.~Roukos, T.~Ward, and W.-J. Zhu, ``{BLEU: A Method for Automatic
  Evaluation of Machine Translation},'' in \emph{ACL}, 2002, pp. 311--318.

\end{thebibliography}

% \begin{thebibliography}{5}
% %
% \bibitem {vin:tosh}
% Vinyals, O., Toshev, A., Bengio, S. and Erhan, D., 2015. Show and tell: A neural image caption generator. In Proceedings of the IEEE conference on computer vision and pattern recognition (pp. 3156-3164).

% \bibitem {zhang:liu:ram:zhug}
% Zhang, X., Liu, X., Ramachandran, A., Zhuge, C., Tang, S., Ouyang, P., Cheng, Z., Rupnow, K. and Chen, D., 2017, September. High-performance video content recognition with long-term recurrent convolutional network for FPGA. In Field Programmable Logic and Applications (FPL), 2017 27th International Conference on (pp. 1-4). IEEE.

% \bibitem {chang:mart}
% Chang, A.X.M., Martini, B. and Culurciello, E., 2015. Recurrent neural networks hardware implementation on FPGA. arXiv preprint arXiv:1511.05552.

% \bibitem {li:wu}
% Li, S., Wu, C., Li, H., Li, B., Wang, Y. and Qiu, Q., 2015, May. Fpga acceleration of recurrent neural network based language model. In Field-Programmable Custom Computing Machines (FCCM), 2015 IEEE 23rd Annual International Symposium on (pp. 111-118). IEEE.

% \bibitem{zhang:li:sun}
% Zhang, C., Li, P., Sun, G., Guan, Y., Xiao, B. and Cong, J., 2015, February. Optimizing fpga-based accelerator design for deep convolutional neural networks. In Proceedings of the 2015 ACM/SIGDA International Symposium on Field-Programmable Gate Arrays (pp. 161-170). ACM.

% \end{thebibliography}

\clearpage
\addtocmark[2]{Author Index} % additional numbered TOC entry
\renewcommand{\indexname}{Author Index}
\printindex
\clearpage
\addtocmark[2]{Subject Index} % additional numbered TOC entry
\markboth{Subject Index}{Subject Index}
\renewcommand{\indexname}{Subject Index}
\end{document}